\documentclass[runningheads]{llncs}

\usepackage{microtype}
\usepackage{graphicx}
\usepackage{subfigure}
\usepackage{booktabs} 
\usepackage{hyperref}
\usepackage{natbib}

\usepackage{amsmath}
\usepackage{mathtools}
\usepackage{thmtools}

\usepackage{authblk}

\usepackage[group-separator={,},binary-units=true]{siunitx} 
\usepackage{multirow, bigdelim}

\newcommand{\defd}{\overset{\text{def}}{=}}

\newcommand{\pder}[2][]{\frac{\partial#1}{\partial#2}}
\newcommand{\pderline}[2][]{\partial#1 / \partial#2}

\newcommand{\todo}[1]{}
\renewcommand{\todo}[1]{{\textcolor{red}{ TODO: {#1}}}}

\DeclarePairedDelimiterX{\KLx}[2]{(}{)}{%
  #1\,\delimsize\|\,#2%
}
\newcommand{\KL}{\text{D}\KLx}

\newcommand{\vect}[1]{\boldsymbol{\mathbf{#1}}}

\DeclarePairedDelimiterX{\BDx}[2]{(}{)}{%
  #1,#2%
}

\DeclareDocumentCommand{\comp}{o m o}
{%
  \IfValueT{#1}{#2_{#1}}\IfValueF{#1}{\vect{#2}}\IfValueT{#3}{^{(#3)}}
}

\newcommand{\modloss}{L_{\lambda}}

\newcommand{\etahatj}{\hat{\eta}_j}

\newcommand{\etabar}{\bar{\eta}}

\let\oldtextbf=\textbf
\renewcommand\textbf[1]{{\boldmath\oldtextbf{#1}}}
\newcommand{\mlpone}{$1\times 1024$H}
\newcommand{\mlptwo}{$16\times 64$H}
\newcommand{\mlpthree}{$64\times 16$H}
\newcommand{\mlpfour}{$256\times 4$H}
\newcommand{\dnhigh}{DN-High}
\newcommand{\dnmid}{DN-Mid}
\newcommand{\dnlow}{DN-Low}

\usepackage{hyperref}
\usepackage[capitalise,noabbrev]{cleveref}

\begin{document}
\title{To Ensemble or Not Ensemble:\\ When does End-To-End Training Fail?}%
%
\author{
Andrew Webb\inst{1} \and
Charles Reynolds\inst{1} \and
Wenlin Chen\inst{1} \and
Henry Reeve\inst{2} \and
Dan Iliescu\inst{3} \and
Mikel Luj\'an\inst{1} \and
Gavin Brown\inst{1} }
\authorrunning{Webb et al.}
%
\institute{University of Manchester, UK \and
University of Bristol, UK \and
University of Cambridge, UK\\
}
\maketitle              

\begin{abstract}

%

End-to-End training (E2E) is becoming more and more popular to train complex Deep Network architectures.
An interesting question is whether this trend will continue---are there any clear failure cases for E2E training?
%
%
We study this question in depth, for the specific case of E2E training an {\em ensemble} of networks. 
Our strategy is to blend the gradient smoothly in between two extremes: from independent training of the networks, up to to full E2E training.  
We find clear failure cases, where overparameterized models {\em cannot be trained E2E}.  A surprising result is that the optimum can sometimes lie in between the two, neither an ensemble or an E2E system. The work also uncovers links to Dropout, and raises questions around the nature of ensemble diversity and multi-branch networks.

%

\end{abstract}

\section{Introduction}

Ensembles of neural networks are a common sight in the Deep Learning literature, often at the top of Kaggle leaderboards, and a key ingredient of now classic results, e.g. AlexNet \citep{Alex2012ImageNet}.
In recent literature, an equally common sight is {\em End-to-End} (E2E) training of a deep learning architecture, using a single loss  to train all components simultaneously.  
An interesting scientific question is whether the E2E paradigm has limits, examined in some detail by \citet{glasmachers2017limits}, who concluded that E2E can be inefficient, and {\em ``does not make optimal use of the modular design of present neural networks''}.

In line with this question, some have explored training an {\em ensemble} with E2E, as if it was a modular, multi-branch architecture. 
\citet{dutt2020coupled} explore E2E ensemble training and demonstrate it allows one to \emph{``significantly reduce the number of parameters''} while maintaining accuracy.
\citet{anastasopoulos2018leveraging} use the principle to improve accuracy in multi-source language translation.
\cite{furlanello2018born} show good performance across a variety of standard benchmarks.
This may seem promising, however, \citet{lee2015m} conclude the opposite, that such End-to-End training of an ensemble is {\em harmful} to generalization accuracy.
The idea raises interesting scientific questions about learning in ensemble/modular architectures. 
Some authors have noted that some architectures (e.g. with multiple branches, or skip connections) may be viewed as an ensemble of jointly trained sub-networks, e.g. ResNets \citep{veit2016residual,zhao2016on}. 
It is widely accepted that ``diversity'' of {\em independently} trained ensembles is beneficial \citep{dietterich2000ensemble,kuncheva2003measures}. and hard evidence lies in the success of the many published variants of Random Forests and Bagging \citep{Breiman1996Bagging}.
But if we train a set of networks {\em End-to-End}, they share a loss function---so their parameters are strongly correlated---seemingly the opposite of diversity.  
There is clearly a subtle relation here, raising hard questions, e.g., what is the meaning of `diversity' in E2E ensembles?

Our aim is to understand when each scheme---{\em Independent} ensemble training, or {\em End-to-End} ensemble training---is appropriate, and to shed light on previously reported results.
%
%
\begin{figure}[ht]
    \centering
    {\includegraphics[width=4cm]{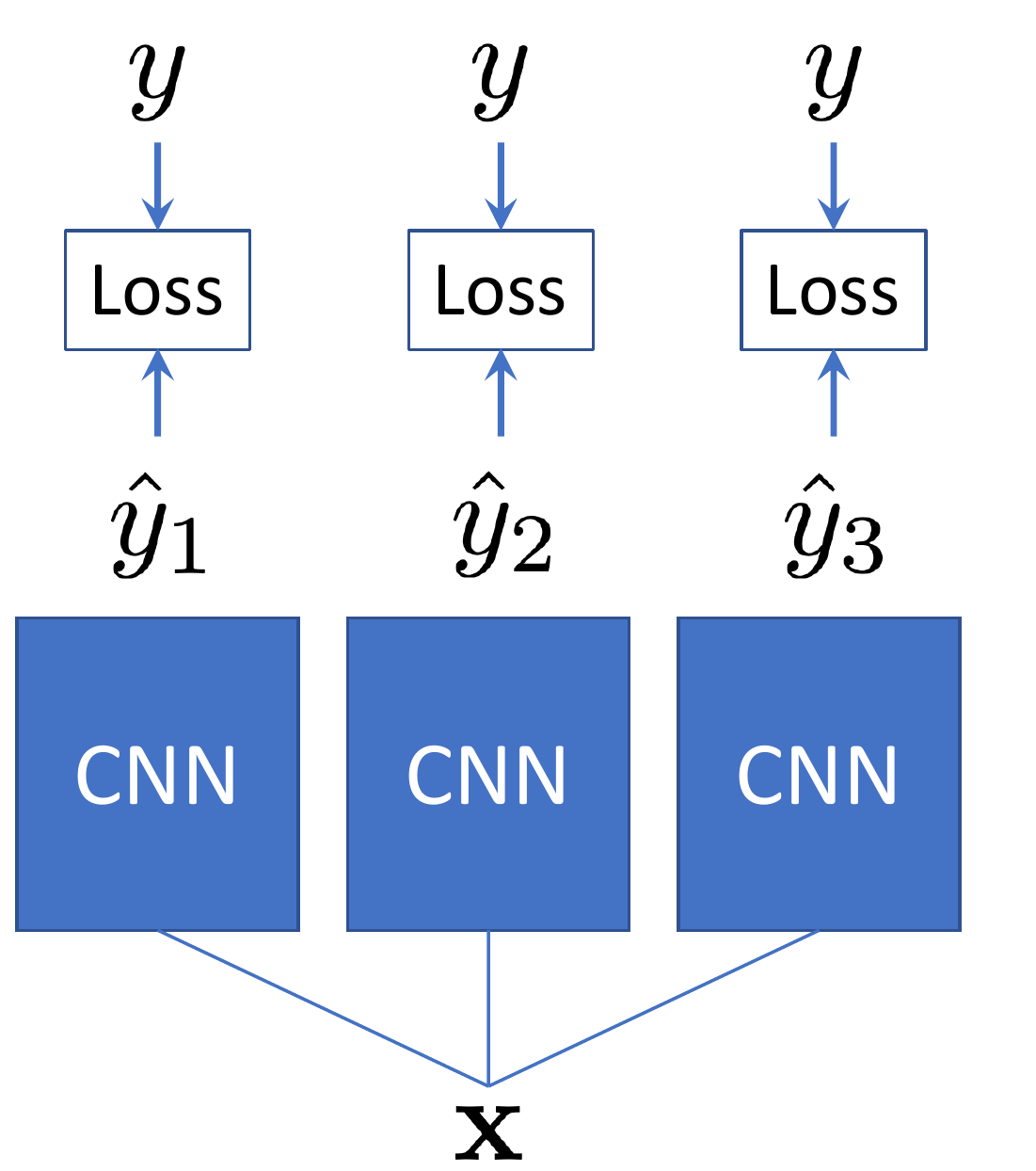}}~~~~~~~~
    {\includegraphics[width=4cm]{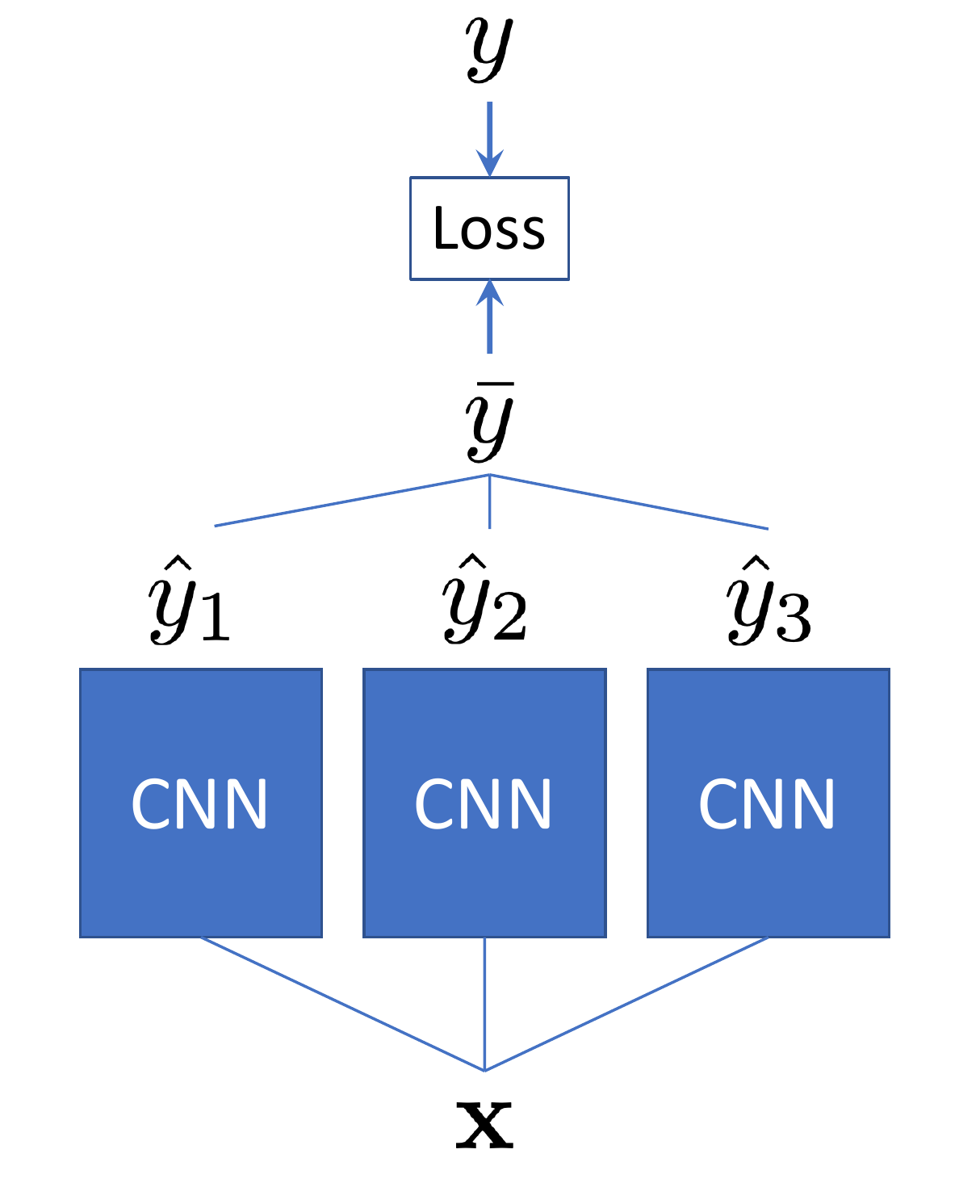}}
    \caption{Computational graphs representing the training of a CNN ensemble: Independent (Left) versus End-to-End training (Right). We study the dynamics of learning when interpolating smoothly in-between these two extremes.}
      \label{fig:schematic}
\end{figure}%

Our strategy is to interpolate the gradient smoothly between the two---a convex combination of Independent and End-to-End training. 
This uncovers complex dynamics, highlighting a tension between individual model capacity and diversity, as well as generating interesting properties, including ensemble robustness with respect to faults.\\

{\noindent \bf Organisation of the Paper:}\\

In Section 2 we cover the necessary notation and probabilistic view we adopt for the paper. In Section 3 we ask and expand upon the reasoning behind the primary question of the paper: {\em is it an ensemble of models, or one big model?}.  This leads us to define a hybrid training scheme, which turns out to generalise some previous work in ensemble diversity training schemes.  In Sections 4 and 5 we thoroughly investigate the question, uncovering a rich source of unexpectedly complex dynamics. Finally in Sections 6 and 7 we outline how this connects to a wide range of work already published, and opens new doors for research.

\newpage
\section{Background and Design Choices}

\label{sec:single-models-and-ensembles}

We assume a standard supervised learning scenario, with training set $S = \{(\vect{x}_i, y_i)\}_{i=1}^n$ drawn i.i.d. from $P(\mathcal{X} \mathcal{Y})$.  
For each observation ${\bf x}$, we assume the corresponding $y$ is a class label drawn from a true distribution $p(y | {\bf x})$.  We approximate this distribution with a model $q(y | {\bf x})$, corresponding to some deep network architecture.  We will explore a variety of architectures: including simple MLPs, medium/large scale CNNs, and finally DenseNets \citep{huang2018multiscaleICLR}. 

To have an ensemble, there is a choice to make in how to combine the network outputs.   In our work we choose to average the network logits, and re-normalise with softmax. This was used by \citet{hinton2015distilling}, and \citet{dutt2020coupled}---we replicate one of their experiments later.   There are of course alternatives, e.g. majority voting or averaging probability estimates.  Majority voting rules out the possibility of E2E ensemble training by gradient methods, since the vote operation is inherently non-differentiable.   Averaging probability estimates is common, but we present a result below that encouraged us to consider the averaging logits approach instead, beyond simply replicating \citet{dutt2020coupled}.

We approach the classification problem from a probabilistic viewpoint: minimising the KL-divergence from a model $q$ to the target distribution $p$.
Given a set of such probability models, we could ask, what is the optimal combiner rule that preserves the contribution from each, in the sense of minimising KL divergence?  In a formal statement we refer to this as the `central' model, denoted $\bar{q}$, that lies at the centre of the set of probability estimates:
\begin{eqnarray}
        {\bar{q}} &=&\arg\min_{z}\Big[ \frac{1}{M}\sum_{i=1}^M D(z\mid\mid q_i)\Big] ~=~ \arg\min_{z}\Big[ \frac{1}{M}\sum_{i=1}^M \int z(y| {\bf x}) \log \frac{z(y| {\bf x})}{q_i(y| {\bf x})} ~d y\Big] 
\end{eqnarray}
It can easily be proved that the minimizer here is the normalized geometric mean, corresponding to a Product of Experts model, though modeling only the means (i.e. a PoE of Generalized Linear Models):
\begin{equation}
\bar{q}(y \mid \vect{x}) \defd Z^{-1} \prod_{j=1}^M q_i(y \mid{\bf x})^{1/M}. \label{poe}
\end{equation}

%
%
This can be written in terms of the distribution's canonical link  $f$ and its inverse $f^{-1}$.  The inverse link for the Categorical distribution is the softmax $f^{-1}(\vect{\eta})$, where $\vect{\eta}$ is a vector of logits.  Correspondingly, the logits are given by the link applied to class probabilities $\vect{\eta}=f(q(y|{\bf x}))$.

\begin{equation}
\bar{q}(y|{\bf x})=f^{-1}\Big(\frac{1}{M}\sum_i f(q_i(y|{\bf x})) \Big)    
\end{equation}
i.e. a softmax operation on the averaged logits --- this provides additional motivation for the ensemble combination rule used by \citet{hinton2015distilling} and \citet{dutt2020coupled}, since it is the rule that preserves the most information from the individual models, in the sense of KL divergence.

\newpage

\section{Is it an Ensemble, or one Big Model?}
\label{sec:modular}

Ensemble methods are a well-established part of the ML landscape.
A traditional explanation for the success of ensembles is the `diversity' of their predictions, and particularly their errors.  When we reach the limit of what a single model can do, we create an ensemble of such models that exhibit `diverse' errors, and combine their outputs. This {\em diversity} can lead to the individual errors being averaged out, and overall lower ensemble error obtained. 

A common heuristic is to have the individual models intentionally lower capacity than they might be, and compensate via diversity. Traditional ensemble methods, such as Bagging and Random Forests, generate diversity via randomization heuristics such as feeding different bootstraps of training data to each model~\citep{Brown2005Diversity}.  {\em Stacking} \citep{wolpert1992stacked} is similar in spirit to E2E ensemble training, in that it trains the combiner function, although only after individual models are fixed, using them as meta-features.

However, if the ensemble combining procedure was fully {\em differentiable}, we could in theory train all networks End-to-End (E2E), as if they were branches or components of one ``big model''.  In this case, what role is there for {\em diversity}?
Furthermore, with modern deep networks, it is easy to have extremely high capacity models, but regularised to avoid overfitting.  With this in mind, is there much benefit to ensembling deep networks?  Various empirical successes, e.g. \citet{Alex2012ImageNet}, suggest a tentative ``yes'', but understanding the overfitting behaviour of deep network architectures is one of the most complex open challenges in the field today.  It is now common to heavily {\em over-parameterize} and regularize deep networks.
These observations raise interesting questions on the benefits of such architectures, and for ensemble diversity.

To study the question of failure cases for E2E ensembles, we could simply train a system E2E and report outcomes. However, to get a more detailed picture, we define a hybrid loss function, {\em interpolating between} the likelihood for an independent ensemble and the E2E ensemble likelihood.
We refer to this as the {\em Joint Training} loss, since it treats the networks jointly as components of a single (larger) network, or as members of an ensemble.  We stress that we are not advocating this as a means to achieve SOTA results, but merely as a forensic tool to understand the behaviour of the E2E paradigm.\\

{\noindent \bf Definition 1} ({\em Joint Training Loss}):
%
\begin{align}
L_\lambda &\defd \lambda \KL{p}{\bar{q}} + (1-\lambda) \frac{1}{M} \sum_{j=1}^M \KL{p}{q_j} \, , \label{eq:convex_combination_form}
\end{align}
where $D$ is the KL-divergence as defined before. 
The loss is a convex combination of two extremes: $\lambda=1$, where we train
$\bar{q}$
as one system, and $\lambda=0$, where we train the ensemble independently (with a learning rate scaled by $1/M$).
%
%
When $\lambda$ lies between the two, it could be seen as $\bar{q}$ being `regularized' by the individual networks partially fitting the data themselves.  Alternatively, we can view the same architecture as an {\em ensemble}, but trained {\em interactively}.  This is made clear by rearranging Equation \eqref{eq:convex_combination_form} as:
\begin{align}
L_\lambda &= \frac{1}{M} \sum_{j=1}^M \KL{p}{q_j} - \frac{\lambda}{M} \sum_{j=1}^M \KL{\bar{q}}{q_j} \label{eq:modular_loss_diversity} ~ .
\end{align}
A simple proof is available in supplementary material.
This alternative view shows that $\lambda$ controls a balance between the average of the individual losses, and a term
measuring the {\em diversity between the ensemble members.} 

From this perspective, this is a diversity-forcing training scheme similar to previous work \citet{Liu1999Ensemble}. In fact for the special case of a Gaussian $p,q$, Eq \eqref{eq:modular_loss_diversity} is {\em exactly} that presented by \citet{Liu1999Ensemble}. However, the probabilistic view allows us to generalise, and for the case of classification problems, this can be seen as {\em managing the diversity in a classification ensemble.}
However, seen as a `hybrid' loss, \eqref{eq:convex_combination_form}, explicitly varies the gradient in-between the two extremes of an ensemble and a single multi-branch architecture, thus including architectures studied previously \citep{lee2015m, dutt2020coupled} as special cases.
In the following section we use this to study the learning dynamics in terms of tensions between model capacity and diversity.




\section{Experiments}


%
%

End-to-End training of ensembles raises several interesting questions:
How does End-to-End training behave when varying the capacity/size of networks?
Should we have a large number of simple networks? Or the opposite---a small number of complex networks?
What effect does varying $\lambda$ have in this setting, smoothly varying from Independent to End-to-End training?
Suppose we have a high capacity, well-tuned {\em single} network, performing close to SOTA.
In this situation, presumably, independent training of an ensemble of such models will still {\em marginally} improve over a single small model, due to variance reduction.   However, it is much less clear in this case whether there will be further gains as $\lambda\rightarrow1$, toward End-to-End training.
We investigate this
first with simple MLPs, exploring trade-offs while the number of learnt parameters is held constant, and then with higher capacity CNNs, including DenseNets \citep{huang2017densely}.

%
%
%

%

\label{sec:capacity-trade-off}

\paragraph{Spending a Fixed Parameter Budget.}
We first compare a single large network to an ensemble of very small networks: each with the {\em same} number of parameters.
The study of such architectures may have implications for IoT/Edge compute scenarios, with memory/power constraints.  We set a budget of memory or number of parameters, and ask how best to ``spend'' them in different architectures.  This type of trade-off is well recognised as highly relevant in the current energy-conscious research climate, e.g. very recently \citet{zhu2019binary}.
We stress again however  that we are not chasing state of the art performance, and instead observe the dynamics of Independent vs E2E training in a controlled manner.

%
%
%

We use single layer MLPs
in four configurations with ${\sim}\num{815}$K parameters:
a single network with 1024 nodes (\mlpone),
16 networks each with 64 nodes (\mlptwo),
64 networks with 16 nodes (\mlpthree),
and 256 networks with 4 nodes (\mlpfour).
We evaluate these on the 10-class classification problem, Fashion-MNIST.
Full experimental details are in the supplementary material.

\Cref{fig:fashion-mnist-lambda-bar} summarizes results for the 1$\times$1024H network versus the 256$\times$4H ensemble. Again, these have the same number of configurable parameters, just deployed in a different manner.
The `monolithic' 1024 node network (horizontal dashed line, 10.2\% error) significantly outperforms the ensemble trained independently (16.3\%), as well as the `classic' methods of Bagging and Stacking. 
However, {\em End-to-End training} of the small network ensemble almost matches performance, and Joint~Training ($\lambda=0.95$) gets closer with 10.3\% error.

  \begin{figure}[ht]
     \centering
       \includegraphics[width=0.6\textwidth]{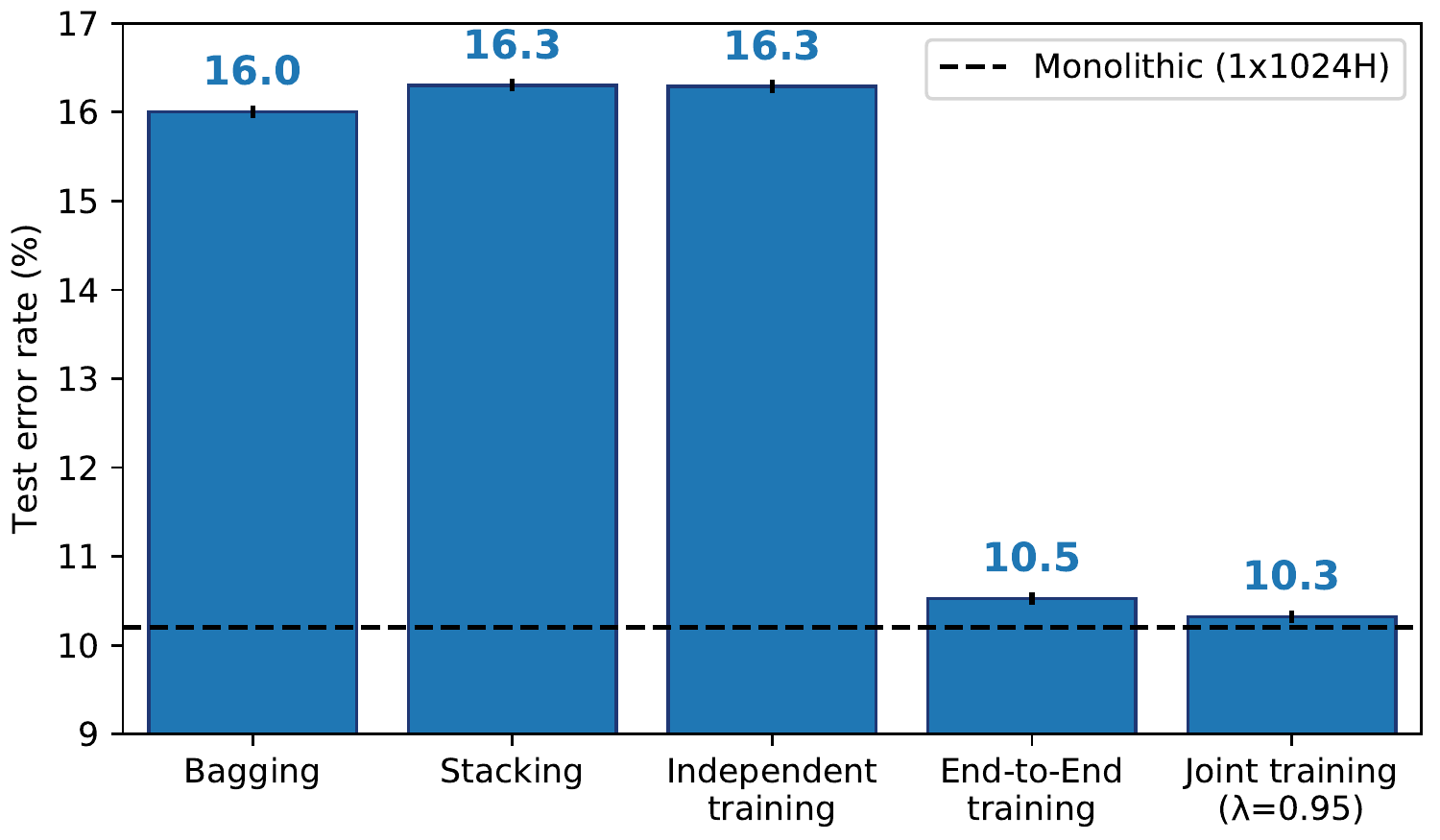}
       \caption{Large single network vs. small net ensemble. Joint Training the ensemble ($\lambda=0.95$)  comes closest to match the large network accuracy, while classic ensemble methods significantly underperform.}
       \label{fig:fashion-mnist-lambda-bar}
 \end{figure}

\begin{figure}[ht]
     \centering
       \includegraphics[width=0.65\textwidth]{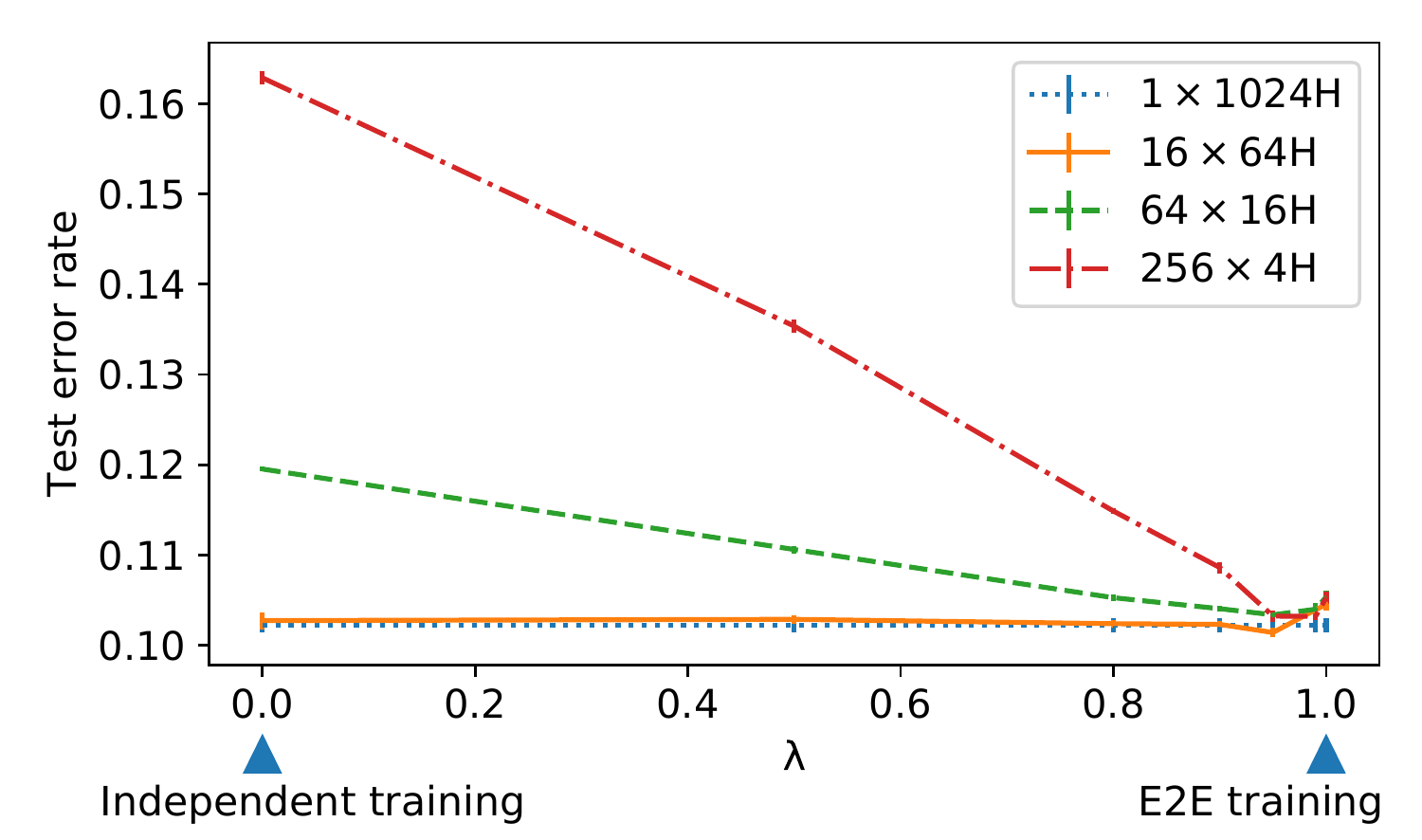}
       \caption{Error rates for MLPs with $815$K parameters, on Fashion-MNIST. Joint Training with higher $\lambda$ values improves the performance of small network ensembles, matching the performance of a single large network.}
       \label{fig:fashion-mnist-lambda}
 \end{figure}

\noindent\Cref{fig:fashion-mnist-lambda} shows detailed results, varying $\lambda$ on other configurations. The distinction between End-to-End training and independent training is most pronounced for the large ensemble of small networks, \mlpfour{},
where {\em E2E training is sub-optimal}. 

However, as the capacity of the networks increases, the difference between End-to-End and independent training vanishes --- illustrated by the almost uniform response to varying $\lambda$ with larger networks.  This is investigated further in the next section with state-of-the-art {\em DenseNet} networks \citep{huang2017densely} with millions of parameters.

\paragraph{High Capacity Individual Models.}
In deep learning, a recent effective strategy is to
massively overparameterize a model and then
rely on the implicit regularizing properties of stochastic gradient descent.
We ask, is there anything to be gained from End-to-End training of such large models?
What does the concept of diversity even mean when the models have almost zero training error?
We address this by examining ensembles of DenseNets~\citep{huang2017densely}, which achieve close to SOTA at the time of writing.
We train a variety of DenseNet ensembles on the CIFAR-100 dataset, with the size of the ensemble increasing as the complexity of each ensemble member decreases, such that each configuration occupies approximately 12GB of GPU RAM. These configurations are described in \Cref{tab:densenet-architecture}. Full experimental details can be found in the supplementary material.

\begin{table}[ht]
\centering
\caption{DenseNet ensembles. A proxy measure of capacity is $(d,k)$, the depth $d$ and growth rate $k$, which we decrease as the size $M$ of the ensemble increases.  Each architecture occupies approximately 12GB RAM.}
\label{tab:densenet-architecture}
\begin{tabular}{@{}lccccc@{}} \toprule
\bf Name & \textbf{Depth} & \textbf{$k$} & \textbf{$M$} & \textbf{Memory} & \textbf{Params.} \\ \midrule
\dnhigh & 100 & 12 & 4 & $\sim$12GB & $3.2$M \\
\dnmid & 82 & 8 & 8 & $\sim$12GB & $2.1$M \\
\dnlow & 64 & 6 & 16 & $\sim$12GB & $1.7$M \\
\bottomrule
\end{tabular}
\end{table}

\Cref{tab:densenet} shows results for independent, End-to-End, and Joint training, vs. Bagging/Stacking.
Each row contains results for a DenseNet ensemble, with minimum error rate in bold.
Results for \dnhigh{} replicate the results of  \citet[Table 2]{dutt2020coupled}.

\begin{table}[ht]
\centering
\caption{Error rates (\%) for DenseNet ensembles. Independent training is optimal (bold) for all but the smallest capacity networks.}
\label{tab:densenet}
\begin{tabular}{@{}lccc@{}} \toprule
~             & \dnlow          & \dnmid          & \dnhigh \\ \midrule
$\lambda=0.0$ {\tiny{(Independent)}} & $25.0$          & $\mathbf{19.9}$ & $\mathbf{17.8}$ \\
$\lambda=0.5$ & $\mathbf{23.5}$ & $20.0$          & $18.8$ \\
$\lambda=0.9$ & $25.9$          & $22.4$          & $20.3$ \\
$\lambda=1.0$ {\tiny{(End-to-End)}}  & $29.6$          & $25.7$          & $22.1$ \\ \midrule
Bagging       & $32.5$          & $29.5$          & $28.4$ \\
Stacking      & $28.3$          & $23.4$          & $21.4$ \\
\bottomrule
\end{tabular}
\end{table}

In the previous section, we saw End-to-End training outperforming independent training for large ensembles of very small models.  Here, {\em we find the opposite is true for small ensembles of large, SOTA models}.  In every DenseNet configuration, {\em End-to-End training (i.e. $\lambda = 1$) is sub-optimal.}
In all but \dnlow{}, which is the configuration with the largest ensemble of smallest models, independent training ($\lambda=0$) achieves the lowest test error.
{\em\bf These results indicate there is little to no benefit in test error when E2E ensemble training SOTA deep neural networks}.   However, the result for \dnlow{} suggests that there may be a relatively smooth transition, somewhere between E2E ensembles and independent training, as the complexity of the ensemble members increases. 


\paragraph{Intermediate capacity.}
%

In this section we further explore the relationship between ensemble member complexity
and E2E ensemble training. We train ensembles of convolutional networks on CIFAR-100. Each ensemble has 16 members of varying complexity, from $70,000$ parameters up to $1.2$ million.
Full details can be found in the supplementary material.  Note that these experiments are not intended to be competitive with the SOTA, but to illustrate relative benefits of E2E ensembles as the complexity of individual members varies.

\begin{figure}[!h]
    \centering
      \includegraphics[width=0.6\textwidth]{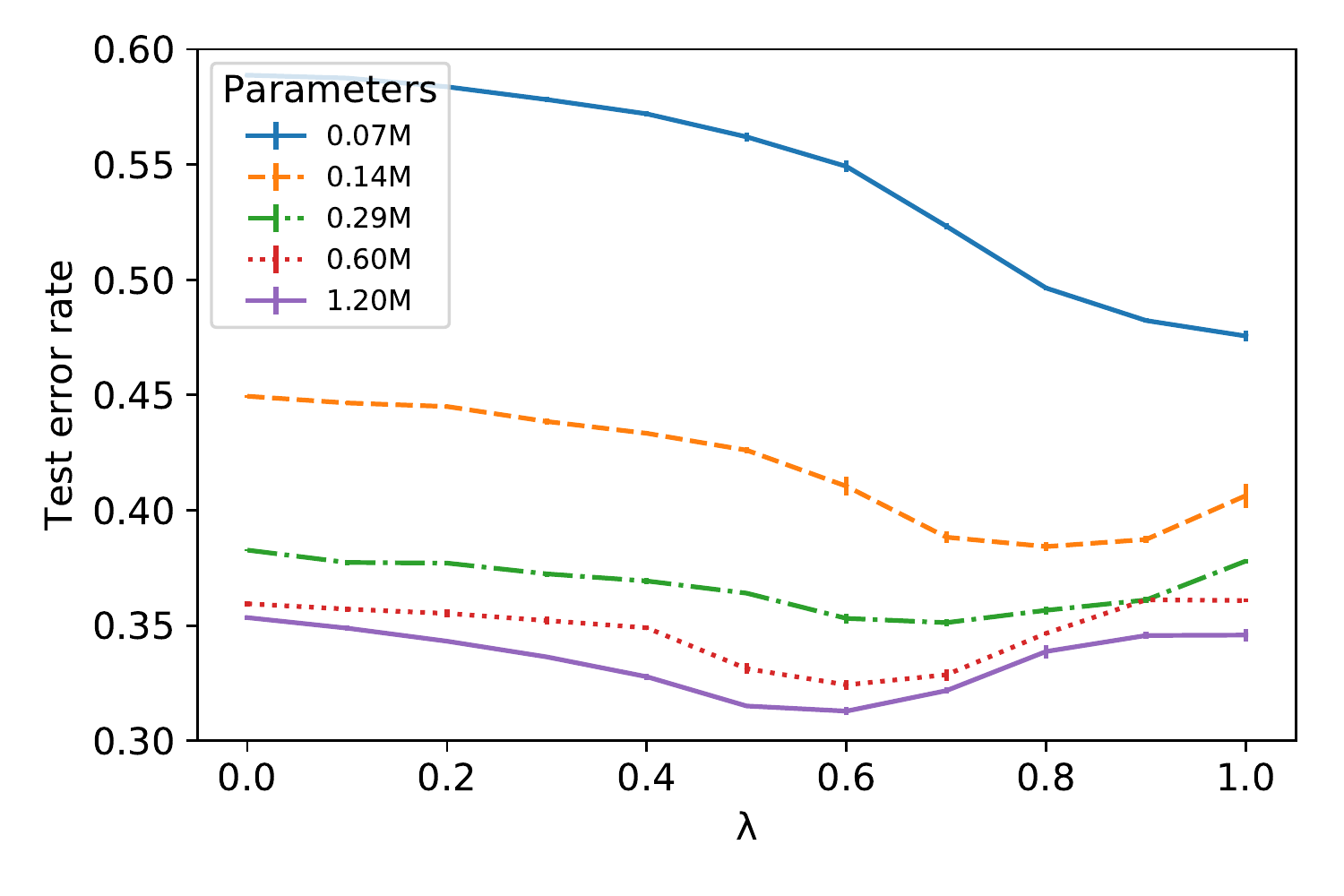}
      \caption{Ensemble test error against $\lambda$ for ensembles of convolutional networks of
      varying complexity on CIFAR-100. The optimal $\lambda$ value shifts away from E2E ensemble training with greater individual network capacity.}
      \label{fig:smaller-convs}
\end{figure}

\Cref{fig:smaller-convs} shows the test error rate and standard errors for each $\lambda$ value for each configuration.
The `dip' in each of the four lines in the figure is the optimal $\lambda$. 
We find that the optimal point smoothly decreases: E2E ensembles are the preferred option with simpler networks, but this changes as the networks become more complex.  At 1.2M parameters, the optimal point lies clearly {\em between} the two extremes.
This supports the trend observed in the previous sections, that the benefits of E2E training as a scheme depend critically on the complexity of individual networks.

\begin{figure}[!h]
	\centering
		\includegraphics[width=10cm]{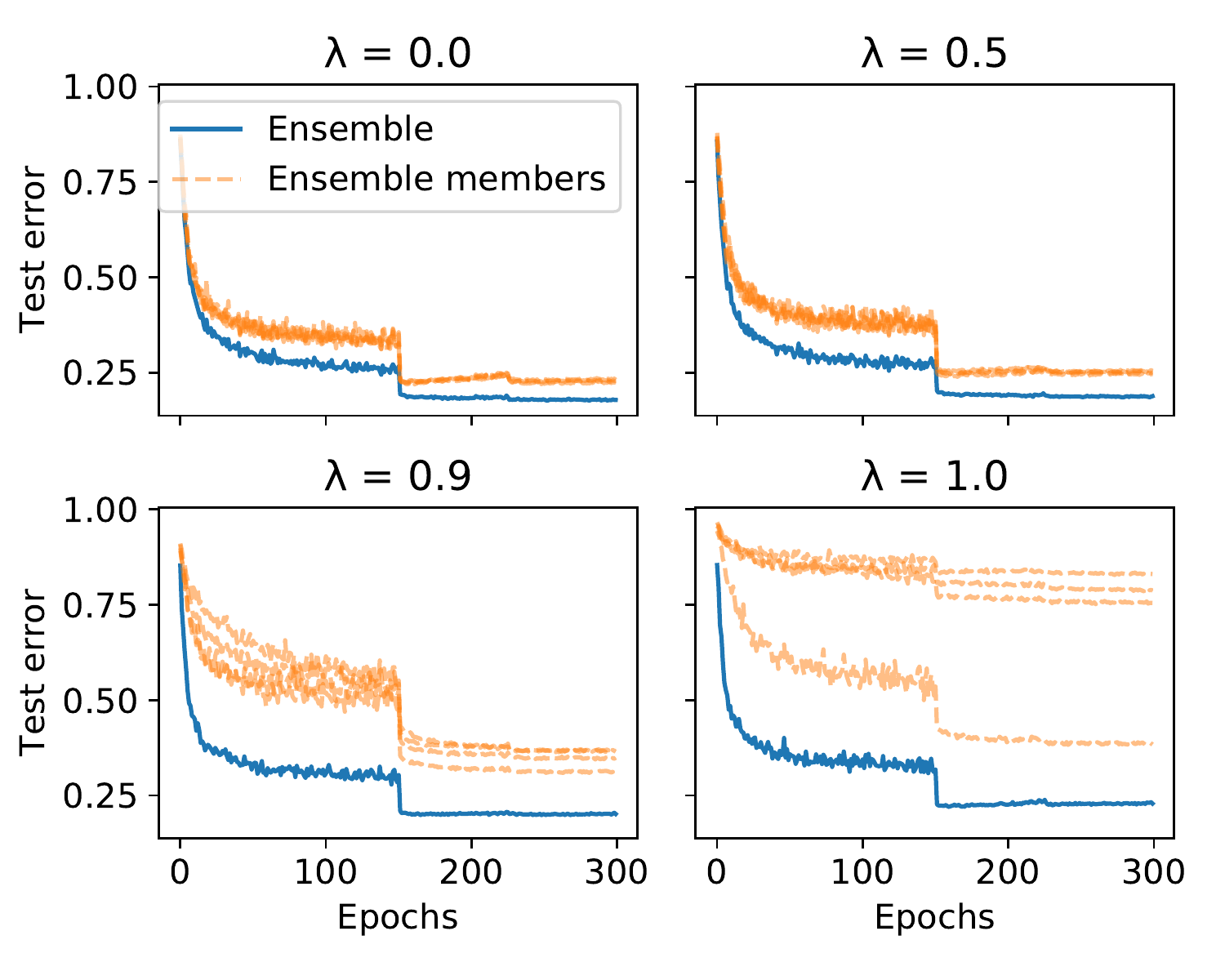}
		\includegraphics[width=10cm]{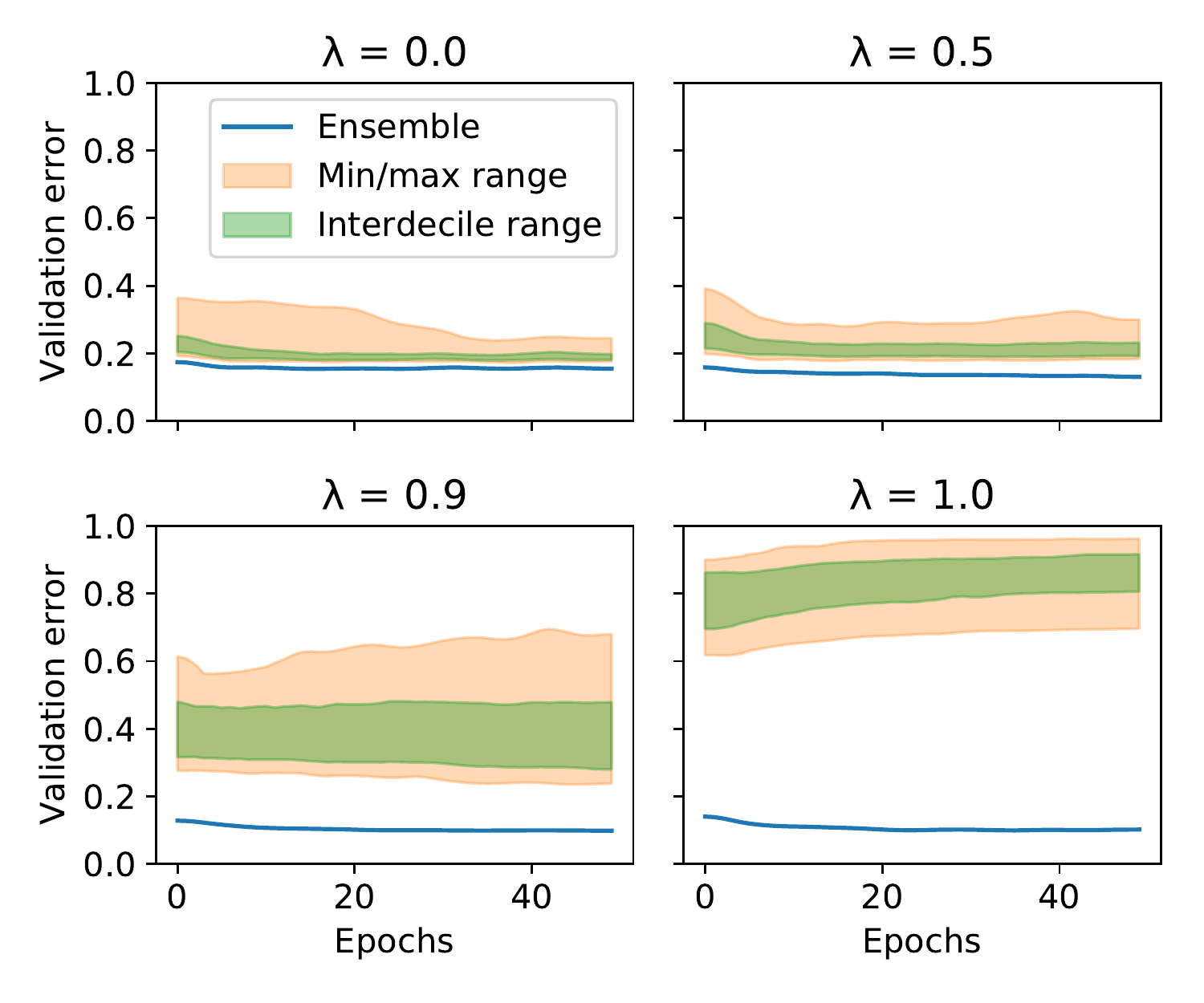}

		\caption{Ensemble vs Member error: Over-capacity models (DenseNets, top sub-figure) experience a `model dominance' effect when trained E2E, whereas there is no such effect with \mlpfour{} undercapacity models (bottom sub-figure).}
		\label{fig:densenet-epochs}
\end{figure}

\paragraph{When does End-to-End ensemble training fail?}
These results suggest a trend:
ensembles of low capacity models
benefit from E2E training, and high capacity models perform better
if trained independently.
We suggest a possible explanation for this trend, and a diagnostic for
determining during training whether E2E training will perform well.
We find that ensembles of high capacity models, when trained E2E, exhibit a
`model dominance' effect, shown in
\Cref{fig:densenet-epochs}.
There is a sharp transition in behaviour at $\lambda=1$ (E2E), whereby a single ensemble member
individually performs well---in both training and test error---while all other members perform poorly. This effect is not observed in
ensembles of low capacity models.

We suggest that this dominance effect can explain the poor performance of E2E training seen in \Cref{tab:densenet}.
We also note that model dominance occurs in the E2E training experiments of \citet[Table 2]{dutt2020coupled} (called `coupled training (FC)' by the authors);
the authors report the average and standard deviation (within a single trial) of the ensemble member error rates. In the case with 2 ensemble members, it can be inferred that one ensemble member achieves a much lower error rate than the other.

\begin{figure}[h]
    \centering
      \includegraphics[width=6cm]{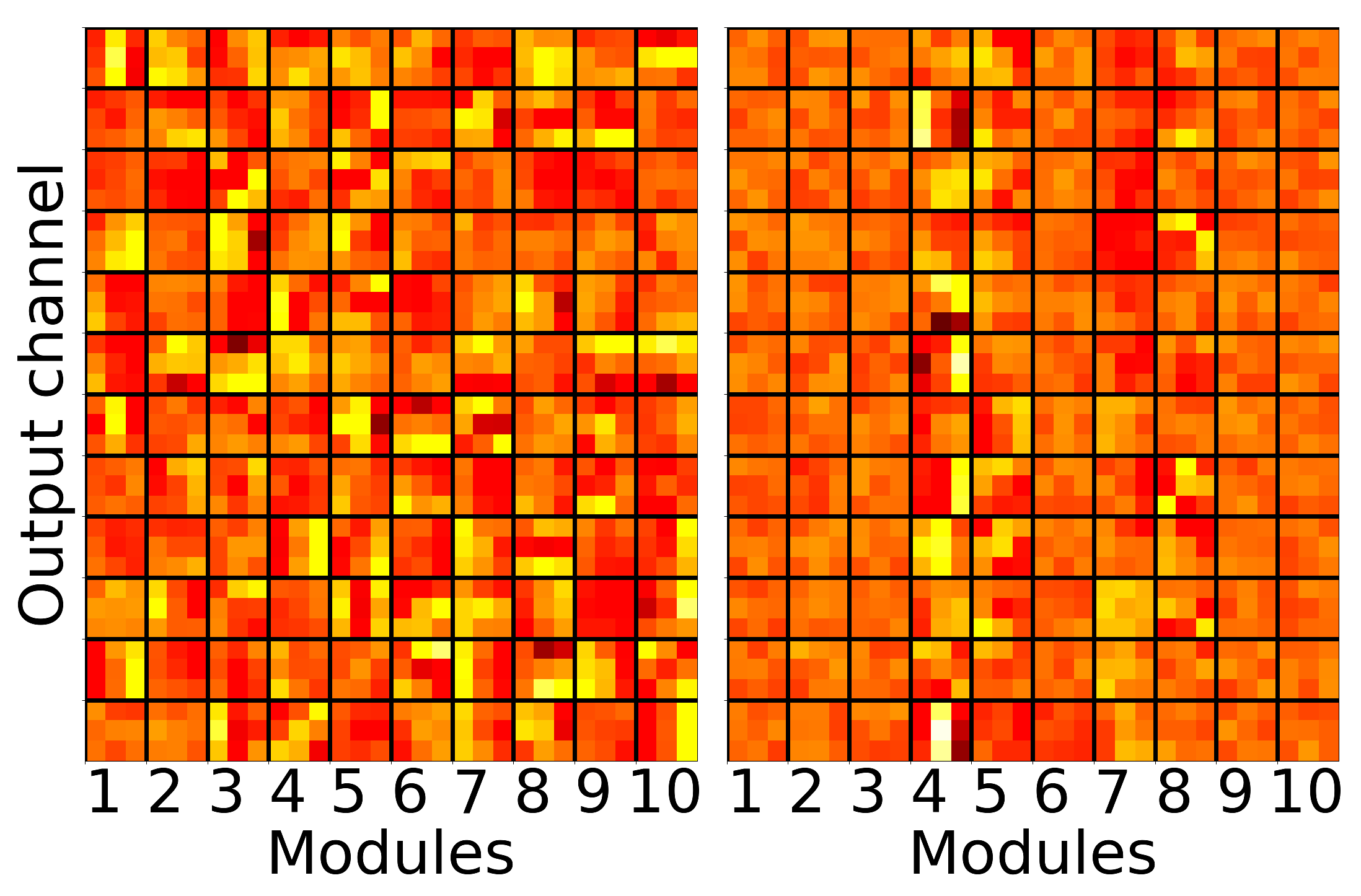}
      \caption{Model dominance in filters in the first layer of 10 CNN modules.  Trained independently (left) and E2E (right). In E2E training, only network 4 has strong filters.}
      \label{fig:filter-heatmap}
\end{figure}

In our own experiments, the dominance effect manifests early in training (\Cref{fig:densenet-epochs}). In each trial, the ensemble member with the lowest error rate by the end of epoch 3 dominates by the end of training; model dominance can be used as an early diagnosis during training that the ensemble members are overcapacity for E2E training.
Note that model dominance can occur in classification because the ensemble prediction is a normalized geometric mean; a network closely fitting the data can arbitrarily reduce the ensemble error---the well-known `veto' effect in Products of Experts~\citep{Welling2007poe}---and the parameters of other networks can be prevented from moving far from their initial values.
This `stagnation' of other ensemble members can be seen in \Cref{fig:filter-heatmap}, which shows some of the filters in the first layer of a ConvNet trained independently and E2E, showing a small number of strong filters in the latter case.


\newpage
\section{What happens {\em between} E2E and independent training?}
\label{sec:observations}


The previous section has demonstrated clearly that E2E training of an ensemble can be sub-optimal. More interestingly perhaps, the complex behaviour seems to lie between independent and E2E.  This section goes beyond just ensemble test error rate, and looks into properties of the networks {\em inside} the ensemble.

\paragraph{Robustness.} We find that Joint Training with $\lambda$ very slightly less than 1 can greatly increase the {\em robustness} of an ensemble, in the sense that
dropping the response from a subset of the networks
at test time does not significantly harm the accuracy.
This {\em partial evaluation}, performing inference for only a subset of the networks, may be useful in resource-limited scenarios, e.g. limited power budgets in Edge/IoT devices.
\begin{figure}[ht]
    \centering
      \includegraphics[width=0.7\textwidth]{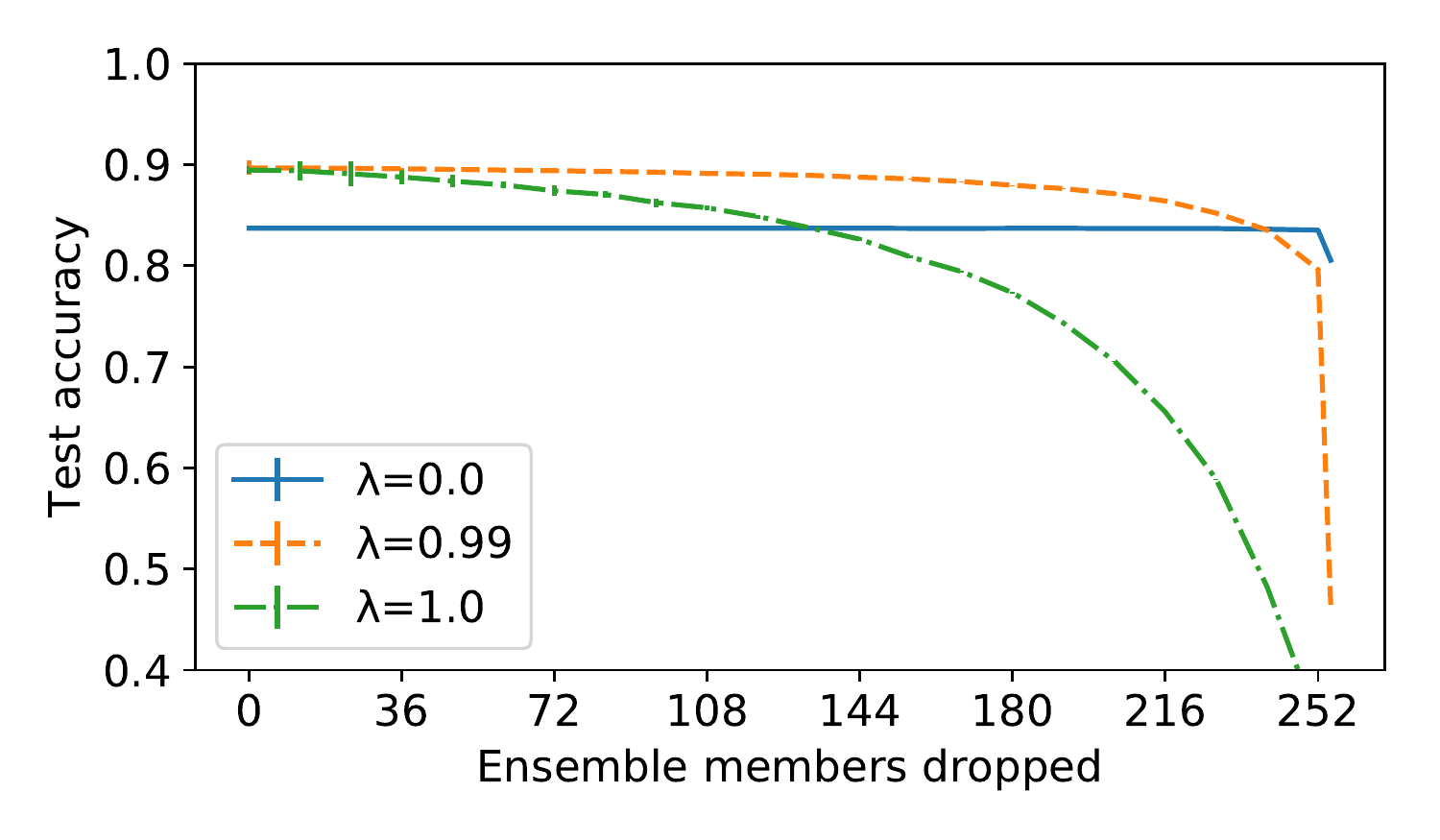}
      \caption{Independent training ($\lambda = 0$) gives robust ensembles; however, many members are redundant. End-to-End training ($\lambda = 1$) gives brittle ensembles. Joint Training with $\lambda\approx 1$ (but strictly less than 1) retains the accuracy whilst adding robustness.}
      \label{fig:fashion-mnist-robustness}
\end{figure}

\begin{figure}[ht]
    \centering
      \includegraphics[width=0.65\textwidth]{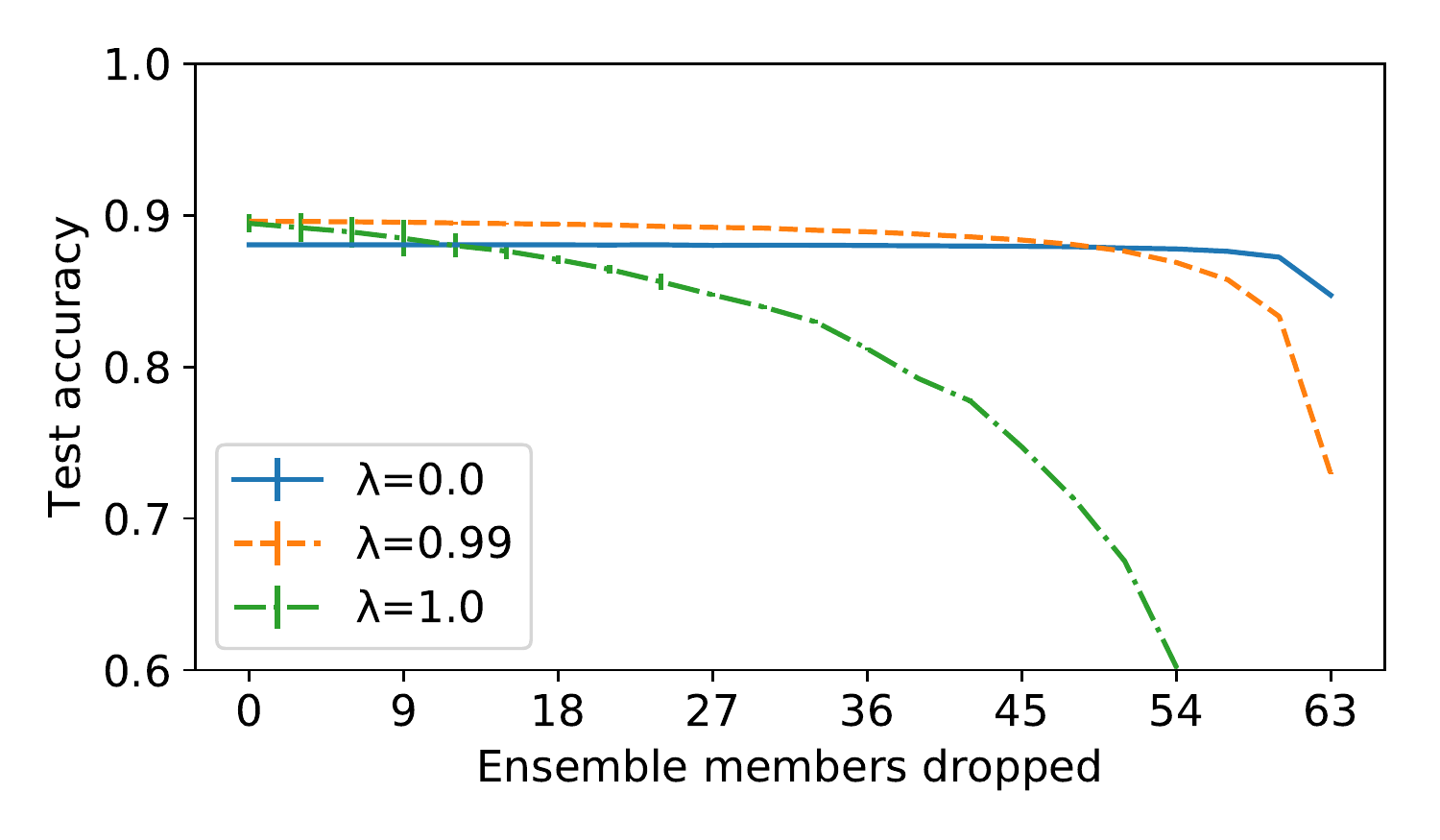}
      \caption{Robustness with the \mlpthree{} architecture. The same robustness appears, but relative benefit is less, as the independently trained ensemble can perform better.}
      \label{fig:fashion-mnist-robustness-64}
\end{figure}

%
\Cref{fig:fashion-mnist-robustness} analyses the \mlpfour{} ensemble, dropping a random subset of networks for each test example in Fashion-MNIST, averaged over 20 repeats.
Independent training ($\lambda = 0$) builds a highly robust ensemble, but which could also be seen to be redundant---adding more networks does not increase accuracy.
End-to-End training ($\lambda = 1$) achieves higher test accuracy, but the ensemble is not at all robust.
Joint Training with $\lambda=0.99$ achieves the same accuracy as End-to-End with all members, but can maintain accuracy even if more than 50\% of the networks are dropped.
%
%
A similar behaviour is observed with the \mlpthree{} architecture, in \Cref{fig:fashion-mnist-robustness-64}.\\

\begin{figure}[ht]
	\centering
		\includegraphics[width=0.65\textwidth]{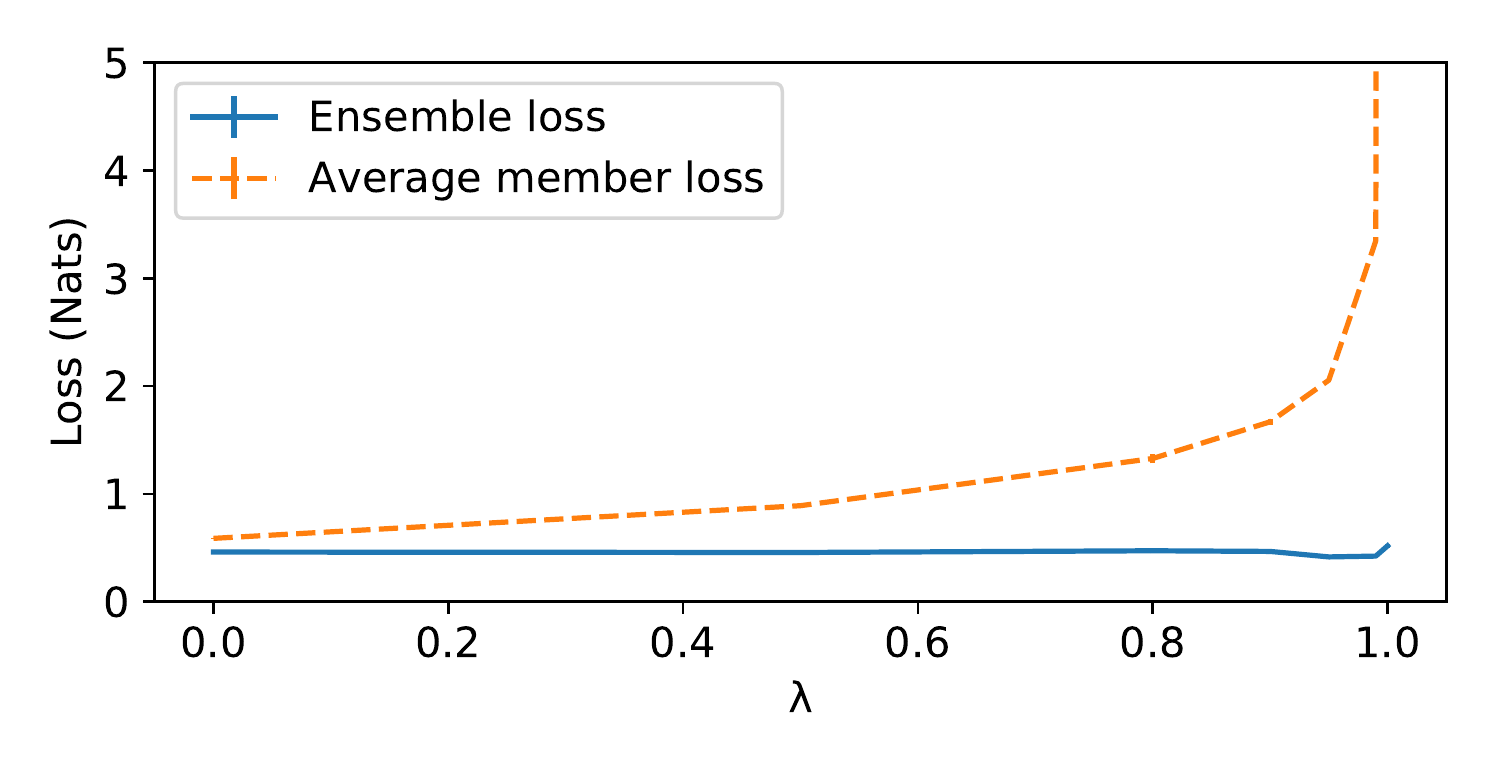}
		\caption{ \mlpthree{}: ensemble vs. average loss. The average grows sharply as $\lambda\rightarrow 1$, a symptom of the individuals relying on others to correct mistakes.}
       \label{fig:fashion-mnist-avg-ens}
\end{figure}

{\bf \em Robust ensembles: why does this occur?} 
When training independently, the networks have no ``idea'' of each others' existence---so solve the problem as individuals---and hence the ensemble is highly robust to removal of networks.
When we {\em End-to-End} train the {\em exact same architecture}, all networks are effectively "slave" components in a larger network. They cannot directly target their own losses, so individual errors would not be expected to be very low.  However, as a system, they work together, and achieve low ensemble error.  The problem is that the networks {\em rely} on the others too much---so when removing networks, the performance degrades rapidly.
This is illustrated in \Cref{fig:fashion-mnist-avg-ens}---we see individual network loss rapidly increasing as we approach End-to-End training (i.e. $\lambda=1$).

The ``sweet spot'' is a small amount of {\em joint training} with $\lambda=0.99$, which ensure the networks cannot rely on each other completely, but are still working together.
%
This is highly reminiscent of {\em Dropout}, 
where one of Hinton's stated motivations was to make each {\em ``hidden unit more robust and drive it towards creating useful
features on its own without relying on other hidden units to correct its mistakes.''} \citep[Section 2]{srivastava2014adropout}.   This similarity is not a coincidence, and is discussed in the next section.


\section{Discussion \& Future Work}
\label{sec:discussion}
The primary question for this paper was {\em ``when does E2E ensemble training fail?''}, with the intention to explain conflicting results in the literature, e.g. \citet{lee2015m, dutt2020coupled}. 
The answer turns out to be intimately tied with the over-parameterization (or not) of individual networks in the ensemble.
\citet{dutt2020coupled,lee2015m} concluded that End-to-End training of ensembles leads to poor performance---however, only very large state-of-the-art models were studied.  Our results agree with theirs {\em exactly}, in situations where ensemble members are over capacity---\Cref{tab:densenet} reproduces the results of \citet[Table 2]{dutt2020coupled}---but, the advantage of E2E emerges with lower capacity individuals, where there are also interesting and potentially useful dynamics in-between Independent/E2E training.
%

{\bf Relations to prior work.}
The investigation turned up links to literature going back over 20 years, in terms of ensemble diversity \citep{Heskes1998Selecting}.
As mentioned, we presented results primarily on classification problems---however, the framework presented in Section 2 applies generally for targets following any exponential family distribution.
For the special case of a Gaussian, and when we re-arrange the loss as \eqref{eq:modular_loss_diversity}, the gradient is {\em exactly} equivalent to a previously proposed method, Negative Correlation Learning (NCL) \citep{Liu1999Ensemble,Brown2005Managing}.
Thus, an alternative view of the loss is as a generalization of NCL to arbitrary exponential family distributions, where the Categorical distribution assumption here can be seen as {\em managing diversity in classification ensembles}, echoing the title of \citet{Brown2005Managing}.
%

{\bf The Dropout connection?}
We observed a qualitative similarity between ensemble robustness, and the motivations behind Dropout \citep{srivastava2014adropout}.
It is interesting to note that Negative Correlation Learning \cite{Liu1999Ensemble} can be proven equivalent to a stochastic dropping of ensemble members during training~\citep{reeve2018modular}, i.e. Dropout at the network level.
With some investigation we have determined that their result depends critically on the Gaussian assumption and hence holds only for NCL, not more generally for Joint Training loss with any exponential family.
Despite this, the observations in \Cref{sec:observations} do suggest a connection, worthy of future work.

{\bf Should we always E2E train Multi-Branch networks? }
The End-to-End training methodology treats the ensemble as if it were a single multi-branch architecture.  A similar view could be taken of {\em any} multi-branch architecture, e.g. ResNeXt \citep{xie2017aggregated} that it is either an ensemble of branches, or a single system.
We have found that if ensemble members are over capacity, then one network can `dominate' an ensemble, leading to poor performance overall.   This raises the obvious open question, of whether it is also true for sub-branches in general branched architectures like ResNeXt.

{\bf Ill-conditioning?}
As $\lambda\rightarrow 1$, the condition number of the Hessian for the JT Loss tends to infinity, as we show in the supplementary material. Therefore there may be cases where E2E training with first-order optimization methods (e.g. simple SGD) performs poorly, and where second-order methods behave markedly differently. We leave this question for future study.

\section{Conclusions}
\label{sec:conclusions}
We have presented a detailed analysis of the question {\em ``When does End-to-End Ensemble training fail?''}.  This is in response to the recent trend of end-to-end training being used more and more in deep learning, and specifically for an ensemble \citep{lee2015m, furlanello2018born, dutt2020coupled}.
Our strategy for this was to study a convex combination of the likelihood-based losses for independent and E2E training, blending the gradient slowly from one to the other. 

Our answer to the question is that End-to-End training tends to under-perform when member networks are significantly over-parameterized, though it is possible to diagnose whether this will happen by examining the relative network training errors in the first few epochs.
In this case we suggest alternative classical methods such as Bagging, or Independent training.  Further, we conjecture that this may have implications for general multi-branch architectures---should we always train them E2E, or perhaps individually?

The space between independent and E2E turned out to be a rich source of quite unexpectedly complex dynamics, generating robust ensembles, with links to Dropout, general multi-branch deep learning, and early literature on ensemble diversity.  We suggest the book is not yet closed and perhaps there is much more to learn in this space.

\section*{Acknowledgements}

The authors gratefully acknowledge the support of
the EPSRC for the LAMBDA project (EP/N035127/1).

\clearpage
\appendix

\section{Re-writing the Joint Training Loss}

Here we show that the convex combination form and the `ambiguity' form of the loss are equivalent. Starting from the convex combination form:
\begin{align}
L_\lambda &\defd \lambda \KL{p}{\bar{q}} + (1-\lambda) \frac{1}{M} \sum_{j=1}^M \KL{p}{q_j} \, , \label{eq:convex_combination_form2}
\end{align}
we use the \emph{ambiguity decomposition}~\citep{Heskes1998Selecting}:
\begin{align}
\KL{p}{\bar{q}} &= \frac{1}{M} \sum_{j=1}^M \KL{p}{q_j} - \frac{1}{M} \sum_{j=1}^M \KL{\bar{q}}{q_j} \label{eq:ambiguity-decomposition} \, .
\end{align}
Substituting the right-hand side for the $\KL{p}{\bar{q}}$ term in \eqref{eq:convex_combination_form2}, we obtain the ambiguity form of the loss:
\begin{align}
L_\lambda &= \frac{1}{M} \sum_{j=1}^M \KL{p}{q_j} - \frac{\lambda}{M} \sum_{j=1}^M \KL{\bar{q}}{q_j} \label{eq:modular_loss_diversity2} \, .
\end{align}

\section{Experimental Details}

Here we specify details such as dataset, model architecture, and training for the modular loss experiments.

\subsection{Spending a Fixed Parameter Budget---MLPs / Fashion-MNIST}

\paragraph{Dataset.}
We use
the Fashion-MNIST dataset \citep{xiao2017fashion}
with the predefined
train/test split, holding out \num{10000} training examples
as a validation set
for early stopping.
We apply mean and standard deviation normalization, and no data augmentation.

\paragraph{Architectures.}

We use single layer MLPs with ReLU activations,
in four configurations each with ${\sim}\num{815}$K parameters:
a single module with 1024 hidden nodes (1-M-1024-H),
16 modules with 64 hidden nodes each (16-M-64-H),
64 modules with 16 nodes (64-M-16-H),
and 256 modules with 4 nodes (256-M-4-H).

\paragraph{Training.}
We train for 200 epochs of SGD, batch size 100, momentum \num{0.9},
and tune the learning rate independently for each configuration and $\lambda$. 
Final reported test error is that at the epoch where validation error was minimized. 
Results are averaged 
over $\num{5}$ trials of random train/validation splits and initializations.

\subsection{High Capacity Individual Models---DenseNets / CIFAR-100}

\paragraph{Dataset.}

We use the
CIFAR-100~\citep{Krizhevsky09learningmultiple} dataset with
the predefined train/test split, per-channel mean and standard deviation normalization, and the standard data augmentation (see, e.g., \citet{He2016}).

\paragraph{Architectures.}

We train ensembles of DenseNet-BC networks~\citep{huang2017densely}. We train 4 modules with a depth of 100 and growth rate 12 (DN-100-12-4)---a configuration used in \citet{dutt2020coupled}. We also train ensembles of 8 and 16 smaller DenseNet modules.
Our DenseNet implementation is based on~\citet{Amos2017repo}. 

\begin{table}[ht]
\centering
\caption{DenseNet architectures.}
\label{tab:densenet-architecture2}
\begin{tabular}{@{}lcccc@{}} \toprule
Name & Depth & $k$ & Modules & Parameters \\ \midrule
\dnhigh & 100 & 12 & 4 & 3.2M \\
\dnmid & 82 & 8 & 8 & 2.1M \\
\dnlow & 64 & 6 & 16 & 1.7M \\
\bottomrule
\end{tabular}
\end{table}

\paragraph{Training.}

We evaluate $\lambda$ values \{0.0, 0.5, 0.9, 1.0\} over 3 trials of parameter initialization.
We use the training procedure described by \citet{huang2017densely,dutt2020coupled}. We use SGD with batch size \num{64}.
The initial learning rate of $0.1$ is decreased by a factor of \num{10} at epochs \num{150} and \num{225}, with momentum \num{0.9}.

\subsection{Intermediate Capacity---Small ConvNets / CIFAR-100}

\paragraph{Dataset.}
We use
the CIFAR-100 dataset~\citep{Krizhevsky09learningmultiple}
with the predefined train/test split, holding out \num{10000} training examples for early stopping. We apply per-channel mean and standard deviation normalization, and
apply the standard flip and crop data augmentation used for this dataset.

\paragraph{Architecture.}

We train ensembles of 16 CNNs with ReLU activations. We apply global pooling before the final fully connected layer, in the style of MobileNets~\citep{howard2017mobilenets}. The networks are fully convolutional, and we evaluate a variety of architectures of varying complexity. The architectures are described in \Cref{tab:conv-architecture}.

\begin{table}[ht]
\centering
\caption{ConvNet architectures. Each architecture is fully convolutional.
Each convolution layer has a $3\times 3$ kernel with no dilation. The `Filters' column indicates the number of output features of each layer. Bold indicates a stride of 2 for a layer, otherwise a stride of 1 is used.}
\label{tab:conv-architecture}
\begin{tabular}{@{}ccc@{}} \toprule
Parameters & Layers & Filters \\ \midrule
0.07M & 3 & 32,{\bf 64},{\bf 64}   \\
0.14M & 4 & 32,{\bf 64},64,{\bf 128}   \\
0.29M  & 5 & 32,{\bf 64},64,{\bf 128},128  \\
0.60M & 6 & 32,64,{\bf 64},128,128,{\bf 256}   \\
1.20M & 7 & 32,64,64,{\bf 128},128,256,{\bf 256}  \\
\bottomrule
\end{tabular}
\end{table}


\paragraph{Training.}

We evaluate $\lambda$ values \{0.0, 0.1, 0.2, \ldots, 1.0\} over \num{5} trials of random training/validation splits and parameter initializations.
We use SGD with batch size \num{128}, with learning rate 0.1,
momentum \num{0.9}, and weight decay $10^-4$.
We train for 400 epochs, decaying learning rate by a factor of 10 at epochs 200 and 300, before reporting test error at the epoch at which validation error is minimized.

\section{Effect of \texorpdfstring{$\lambda$}{Lambda} on the Condition Number of the Hessian}
\label{app:condition}

In this section we show that for $\lambda > 0$, any stationary points---in the special case of scalar model outputs---has a Hessian with both positive and negative eigenvalues, and so all stationary points are saddle points. Further, we show that the condition number of the Hessian grows as $\lambda$ tends to $1$ from below.

For a given input pattern, let the target $y$ be distributed according to a single-parameter exponential family distribution with scalar parameter $\eta$. Let $\etahatj$ be the parameter value prediction for the $j$th model of a collection of $M$
models, and let $\etabar = \frac{1}{M}\sum_{j=i}^M \etahatj$ 
be the ensemble prediction. Let $\hat{y}_j = g(\etahatj)$ and $\bar{y} = g(\etabar)$ be the conditional mean estimates of the $j$th model and ensemble model respectively, where $g$ is the canonical inverse link function of the distribution. We have
\begin{align}
\pder[\modloss]{\hat{\eta}_j} = \frac{1}{M} \Big( (1-\lambda) \hat{y}_j + \lambda \bar{y} - y \Big) \, ,
\label{eq:loss-deriv}
\end{align}
where $L_{\lambda}$ is the modular loss, and the entries of the Hessian are given by
\begin{align}
\frac{\partial^2 L_{\lambda}}{\partial \hat{\eta}_i \partial \hat{\eta}_j} = 
  \begin{cases*}
  \frac{1}{M} \left( 1 - \lambda (1 - \frac{1}{M}) \right) \cdot g^{\prime}(\hat{\eta}_i) & if i = j \\
  \frac{\lambda}{M^2} \cdot g^{\prime}(\etabar) & otherwise
  \end{cases*} \, .
\end{align}
From \eqref{eq:loss-deriv}, for $\lambda \neq 0$ any stationary point of the loss must have $\hat{y}_i = \hat{y}_j = \bar{y} = y$, and therefore $c \defd g^{\prime}(\hat{\eta}_i) = g^{\prime}(\etahatj) = g^{\prime}(\etabar) = g^{\prime}(\eta)$ for all $i, j$, and the Hessian takes the form
\begin{align}
H = \begin{bmatrix} 
    q & r & r & \dots & r \\
    r & q & r & \dots & r \\
    r & r & q & \dots & r \\
    \vdots & \vdots & & \ddots & \vdots \\
    r & r & r & \dots & q \\
    \end{bmatrix}
    \quad \label{eq:hessian} \, ,
\end{align}
with diagonal entries
\begin{align}
q = \frac{1}{M} \left( 1 - \lambda \left(1 - \frac{1}{M}\right) \right) \cdot c
\end{align}
and off-diagonal entries
\begin{align}
r = \frac{\lambda}{M^2} \cdot c \, .
\end{align}
This matrix is $H = r\cdot J_M + (q-r)\cdot I_M$, where $J_M$ is the $M\times M$ matrix of ones and $I_M$ is the $M\times M$ identity matrix. The eigenvalues of $J_M$ are $M$ with multiplicity $1$ and $0$ with multiplicity $M-1$.
Therefore, the eigenvalues of $H$ are
\begin{align}
\omega_1 &= q + (M - 1)\cdot r = \frac{c}{M} \\
\omega_2 &= q - r = \frac{c}{M} \cdot (1 - \lambda) \, .
\end{align}
From this, we can see that for $\lambda > 1$ the Hessian at the stationary point has both positive and negative eigenvalues, and therefore the stationary point is a saddle point, therefore the models will diverge.
Moreover, for $1 > \lambda > 0$, the condition number is
\begin{align}
\kappa(H) = \frac{\omega_1}{\omega_2} = \frac{1}{1-\lambda} \, ,
\end{align}
which tends to infinity as $\lambda$ tends to $1$ from below.
This may suggest that optimization may be problematic for $\lambda$ close to $1$, and that we might see significantly different learning behaviour between first- and second-order methods in this regime.

Note that in the case of the Bernoulli distribution, the loss surface with respect to the parameter estimates $\etahatj$ has no stationary points, but a similar argument can be made with limits.

\section{Equivalence of `Coupled Ensembles' Training Methods}
\label{app:coupled}

We prove here that the `LL' and `SM' coupled training methods of \citet{dutt2020coupled} for ensembles of classifiers are actually equivalent to independent training, up to a scaling of learning rate. We demonstrate that here.

Suppose we have a collection of $M$ neural networks for a $K$ class classification problem.
Let $q_k^{(m)}$ denote the $k$th post-softmax output of the $m$th neural network for a given example.
Let $\mathbf{y}$ be the one hot-encoded true label. The cross entropy loss $L^{(m)}$ of the $m$th network is
\begin{align}
L^{(m)} = -\sum_k y_k \log q_k^{(m)} \quad .
\end{align}

The `LL' coupled training method of \citet{dutt2020coupled} has as its loss function $L_{LL}$ the arithmetic mean of the cross entropy loss functions for each network. I.e.,
\begin{align}
L_{LL} = \frac{1}{M} \sum_m L^{(m)} \quad .
\end{align}

It follows from the fact that $\pderline[L^{(m)}]{q_k^{(n)}} = 0$ if $m\neq n$---i.e., that the cross entropy loss of one network does not depend on the output of another---that
\begin{align}
 \pder[L_{LL}]{q_k^{(m)}} = \frac{1}{M} \pder[L^{(m)}]{q_k^{(m)}} \quad.
\end{align}
In words, the gradient of the `LL' loss with respect to a given network output---and therefore the gradient with respect to the network parameters---is the same as when training independently, scaled by a factor $1/M$.

The `SM' coupled training method of \citet{dutt2020coupled} works as follows. First, take the log of the probabilities $q_k^{(m)}$, and then take the arithmetic mean across networks.
The key point here is that the result is not a vector of log probabilities; it is un-normalized.
An inspection of the authors' provided code~\citep{dutt2018repo} shows that, in the `SM' method, this un-normalized log probability vector is given as input to the \texttt{NLLLoss} loss function provided by PyTorch, which expects log probabilities. The result is that the cross entropy loss is applied to the un-normalized probabilities
\begin{align}
\Tilde{q}_k &= \exp \left( \frac{1}{M} \sum_m \log q_k^{(m)} \right) \quad ,
\end{align}
and that, if $\bf y$ is the one hot-encoded true label, the loss function that is effectively used is
\begin{align}
    L_{SM} &= -\sum_k y_k \log \Tilde{q}_k \\
        &= -\sum_k y_k \frac{1}{M} \sum_m \log q_k^{(m)} \\
    &= \frac{1}{M} \sum_m L^{(m)} = L_{LL} \quad.
\end{align}

This suffices to demonstrate that the `LL' and `SM' methods are equivalent to independent training up to a scaling of learning rate. 




\end{document}